\documentclass[10pt]{article}
\usepackage[utf8]{inputenc}
\usepackage{url}
\usepackage{hyperref}
\usepackage{amsmath}
\usepackage{amsfonts}
\usepackage{amssymb}
\usepackage{graphicx}
\usepackage{float}
\usepackage{xcolor}
\usepackage{tikz}
\usetikzlibrary{arrows.meta,positioning,shapes.geometric}
\usepackage[numbers,square,sort&compress]{natbib}
\usepackage[font=small]{caption}

\usepackage{microtype}
\usepackage{setspace}
\usepackage{enumitem}
\setlist{leftmargin=*, topsep=0.4em, itemsep=0.2em, parsep=0em}

\usepackage{abstract}

\usepackage[section]{placeins}

\usepackage{needspace}
\usepackage{etoolbox}

\pretocmd{\section}{\needspace{5\baselineskip}}{}{}
\pretocmd{\subsection}{\needspace{4\baselineskip}}{}{}
\pretocmd{\subsubsection}{\needspace{3\baselineskip}}{}{}

\usepackage{array}
\usepackage{authblk}

\usepackage[a4paper,left=1.50cm, right=1.50cm, top=1.50cm, bottom=1.50cm]{geometry}

\title{Whispering to a Blackbox: Bootstrapping Frozen OCR with Visual Prompts}

\author[1]{Samandar Samandarov}
\author[1]{Nazirjon Ismoiljonov}
\author[1]{Abdullah Sattorov}
\author[2]{Temirlan Sabyrbayev}

\affil[1]{verido.ai\\\texttt{\{samandar,nazirjon,abdullah\}@verido.ai}}
\affil[2]{Duke University\\\texttt{temirlan.sabyrbayev@duke.edu}}

\date{}

\begin{document}
\maketitle

\begin{abstract}
In the landscape of modern machine learning, frozen pre-trained models provide stability and efficiency but often underperform on specific tasks due to mismatched data distributions. This paper introduces the Whisperer, a novel visual prompting framework that learns diffusion-based preprocessors to adapt inputs in pixel space, effectively "whispering" enhancements to frozen downstream models like EasyOCR. By framing the process as behavioral cloning of stochastically discovered improvement policies, our method achieves an 8\% absolute (10.6\% relative) reduction in Character Error Rate (CER) on a challenging dataset of 300k degraded synthetic text images, surpassing hand-engineered baselines such as CLAHE. The key innovation is a four-stage training curriculum that uses behavioral cloning to amplify "lucky" improvements discovered through the stochastic exploration of a partially trained diffusion model. This approach is highly sample-efficient and avoids the pitfalls of traditional reinforcement learning. Crucially, we frame this not as naïve reinforcement learning, but as behavioral cloning of an exploration policy: we stochastically sample intermediate diffusion outputs, select those that improve CER by chance, and then train the model to reproduce them. This bootstrapping curriculum (4 stages over ~60 GPU-hours) amplifies random successes into a systematic strategy. In summary, by whispering to the frozen OCR through its inputs, we improve an imperfect classifier without touching its weights.
\end{abstract}

\section{Introduction}

\subsection{The Great Unfreezing}

The advent of large language models (LLMs) has fundamentally reshaped the landscape of artificial intelligence, introducing a paradigm where immense, pre-trained models can be adapted to a vast array of downstream tasks without altering their core parameters. This process, known as \textbf{prompting}, has emerged as the dominant method for eliciting desired behaviors from these "frozen" models. By carefully crafting textual inputs, or prompts, users can guide the model's internal representations to generate specific outputs, solve complex problems, or engage in nuanced dialogue. The power of this approach lies in its efficiency and accessibility; it circumvents the prohibitive computational cost and data requirements of fine-tuning, allowing for rapid iteration and deployment. For instance, appending "Let's think step by step" to queries has boosted reasoning in models like GPT-3 by up to 15\% on arithmetic and commonsense benchmarks, without altering parameters.

This simple yet profound intervention highlights a core principle: the knowledge and capabilities are already latent within the frozen model's weights. The prompt acts not as a modification of the model itself, but as a key to unlock and direct its existing potential. This approach has proven to be incredibly efficient, circumventing the prohibitive computational and financial costs associated with fine-tuning models with billions of parameters. The prompt, in this context, functions as a librarian in a vast library of knowledge (the frozen model), guiding the model to the correct "shelf" of information to answer a specific query. This method has been extended to more complex reasoning patterns, such as \textbf{"chain-of-thought" prompting}, which encourages the model to break down problems into intermediate, sequential steps, further enhancing its problem-solving capabilities.

This principle extends beyond simple phrases. Chain-of-thought prompting \citep{wei2022cot} elicits intermediate reasoning traces; in-context learning \citep{brown2020fewshot} embeds demonstrations within the prompt; soft prompting \citep{lester2021prompttuning} learns continuous embeddings in the input space. All share a common thread: \textbf{the frozen model's weights are sacred, but its inputs are malleable}. This paradigm shift has democratized access to state-of-the-art AI, allowing researchers and developers to leverage the power of massive models without the need for extensive retraining or specialized hardware.

This evolution demonstrates a clear trajectory towards treating the input space as a primary locus/object of control, a principle that has been overwhelmingly validated in the domain of natural language processing. The success of this approach in NLP naturally raises a compelling question: can a similar paradigm be established for other modalities, such as computer vision, where models are also increasingly large, powerful, and often deployed as frozen, cloud-hosted APIs?

\subsection{The Gap in Vision}

The temptation is to force an analogy: if text prompts operate on tokens, perhaps "visual prompts" should operate on pixels. But the translation is non-trivial. In vision, the dominant adaptation paradigm remains fine-tuning \citep{he2019rethinking} or, more recently, prompt tuning in \textit{embedding space}—learned tokens prepended to patch embeddings. These methods require architectural access, gradient flow through the model's early layers, and a willingness to modify the inference graph. They are not whispering to a black box; they are surgically altering its inner ear.

For truly frozen models—production APIs like Google Vision AI, proprietary OCR systems, or legacy classifiers—such access is impossible. This mirrors the challenges that led to the rise of prompting in NLP. The alternative, until recently, has been a reliance on \textbf{hand-engineered, deterministic preprocessing} pipelines: comprising steps like denoising, sharpening, contrast enhancement, and binarization, which are designed based on human perceptual principles, with the goal of making an image "cleaner" or "easier to read" for a human observer. However, this approach fundamentally misunderstands the nature of the frozen vision model. The model does not "see" the world as humans do; its feature extractors and internal representations are shaped by its own training data and architecture, leading to unique "biases" and "quirks." A preprocessing step that enhances an image for human eyes might inadvertently destroy or distort critical features that the model relies on for accurate prediction. This creates a significant gap: we are preparing the data for a human interpreter, not for the specific, non-human "language" of the frozen model.

Our work begins with the recognition of this gap. We ask: \textbf{Can we prompt a frozen OCR model in its native pixel space, without architectural access, and without the compute ruin of reinforcement learning?} The answer, we discovered, requires rethinking what it means to "prompt."

\subsection{Visual prompting}
To bridge the gap between human-centric preprocessing and the internal workings of a frozen vision model, we propose \textbf{\textit{Visual Prompting}} not as a metaphor but as a formal optimization problem. Proposed concept extends the core principle of prompting from the text domain to the visual domain, but with a critical adaptation. In NLP, prompts operate in the discrete token space of the language model. In vision, the analogous interface is the continuous pixel space of the input image. Therefore, a visual prompt is not a textual instruction but a learned transformation applied directly to the input image's pixels. The transformation is designed to maximize the performance of a frozen downstream model on a specific task, without modifying the model's weights. This is fundamentally different from traditional transfer learning or adapter-based methods, which typically involve adding new layers or modules to the model architecture. Instead, visual prompting operates entirely in the input space, making it a highly flexible and broadly applicable technique.
\newpage 
\textbf{\textit{Interpretation:}}

Given frozen model M: $\mathbb{R}^{(H \times W \times C)} \rightarrow \mathbb{R}^{L}$ (where L is, e.g., the sequence length in OCR), we seek a preprocessor $P_{\theta}: \mathbb{R}^{(H \times W \times C)} \rightarrow \mathbb{R}^{(H \times W \times C)}$ such that:

$\operatorname{argmax}_{\theta} \mathbb{E}_{x}[M(P_{\theta}(x))] \text{ subject to } ||P_{\theta}(x) - x||_{\infty} < \varepsilon$

The \textit{constraint} is critical: the prompt must be \textbf{imperceptible to humans}, operating in an L{$\infty$} ball of radius $\epsilon$ ($\epsilon$=0.1 in practice). This is not an adversarial attack, which seeks to fool the model; it is a \textbf{constructive whisper}, nudging the input into a region of the model's feature space where it is more confident and accurate.

This formulation connects to bi-level optimization \citep{lorraine2018hypernetworks} and meta-learning \citep{finn2017maml}, but with a twist: the inner level (M) is fixed. We are meta-learning on the \textbf{input distribution}, not the parameters. This is a new interface between human and model: not the prompt as text, but the \textbf{prompt as a learned, pixel-space adaptation}.

Recent work has flirted with this idea. Prompt Generation Networks learn image-dependent prompts for ViTs, but they operate on patch embeddings, not raw pixels. Diffusion models have been used for image restoration \citep{li2023diffusionsurvey}, but they optimize PSNR, not downstream utility. Our work is the first to \textbf{bridge the gap}: we use a diffusion-inspired architecture not for generation, but as a \textbf{parameterized policy} for frozen model adaptation.

\subsection{The Plateau (The Villain)} 

The conventional approach to improving the performance of OCR and other vision models on challenging inputs relies on a pipeline of hand-engineered preprocessing filters. For decades, this has meant a ritualistic sequence: denoise with a bilateral filter, enhance contrast with CLAHE, sharpen with unsharp masking, normalize brightness with gamma correction. The underlying assumption is that an image that is clearer and more legible to a human will also be easier for a machine to process. While this assumption holds to a certain extent, it fundamentally overlooks the fact that a neural network's "perception" is shaped by its training data and architecture, leading to a unique set of internal representations and biases that may differ significantly from human vision. This mismatch creates a performance ceiling, a plateau beyond which further application of generic filters yields diminishing returns or even degrades performance. Our experimental results provide a stark illustration of this limitation.  \setlength{\parskip}{0.4em}

Empirical evidence (see table 1) reveals a performance ceiling for model-agnostic preprocessing. On our 300k MJSynth-style dataset (96×304 resolution, with variable fonts, inclinations, blur, noise, and compression), EasyOCR's baseline mean CER is 0.7724 with confidence 0.32. Hand-engineered filters offer incremental improvements: CLAHE\_4 reduces CER to 0.7142 (confidence 0.33), but others like adaptive Gaussian worsen it to 0.8933. This plateau arises because these filters prioritize perceptual metrics (e.g., PSNR, SSIM) over downstream utility, ignoring convolutional sensitivities in models like EasyOCR

The plateau is not empirical happenstance; it is \textbf{theoretical inevitability}. We formalize the \textbf{Perceptual Alignment Ceiling (PAC)}: \setlength{\parskip}{0.2em}
\begin{equation}
\text{PAC}_{\text{perf}}(M, \mathcal{D}) = 
\sup_{P \in \mathcal{P}_{\text{human}}} 
\mathbb{E}_{x \sim \mathcal{D}}\big[ R(M(P(x))) \big]
\end{equation}

\begin{table*}[t]
\centering
\setlength{\tabcolsep}{0.2cm}
\begin{tabular}{|>{\centering}m{\dimexpr0.15\linewidth-1\tabcolsep}|c|>{\centering}m{\dimexpr0.1\linewidth-1\tabcolsep}|c|>{\centering}m{\dimexpr0.2\linewidth-1\tabcolsep}|c|>{\centering\arraybackslash}m{\dimexpr0.1\linewidth-2}|}

\hline
\textbf{Filter Name} & \textbf{Mean CER} & \textbf{Mean Confidence} & \textbf{Description and Rationale} \\
\hline
Original & \underline{0.7724} & 0.32 & Unprocessed degraded images; baseline for all comparisons. \\
\hline
$CLAHE_{4}$ & \textbf{0.7142} & 0.33 & Contrast Limited Adaptive Histogram Equalization; \\
\hline
$CLAHE_{8}$ & 0.7281 & 0.31 & Higher clip limit (8); trades detail for broader equalization. \\
\hline
$CLAHE_{2}$ & 0.7323 & 0.35 & Lower clip (2); preserves more local contrast. \\
\hline
$Morph Open^{3\times3}$    & 0.7624   & 0.33            & Morphological opening; removes small noise but can erode text edges. \\
\hline

Gamma 1.5         & 0.7689   & 0.31            & Brightens mid-tones; useful for low-light degradations.  \\
\hline                                             
Gamma 1.3         & 0.7732   & 0.33            & Milder gamma; subtle adjustment. \\
\hline                                                                     
Unsharp 0.5       & 0.7816   & 0.32            & Light sharpening; enhances edges without artifacts.         \\
\hline                                          
Unsharp 1.0       & 0.7828   & 0.31            & Standard unsharp mask; balances sharpness and noise.            \\
\hline                                      
Bilateral 30      & 0.7846   & 0.31            & Edge-preserving smoothing; reduces noise while keeping text sharp.       \\
\hline                             
Gamma 0.7         & 0.7848   & 0.30            & Darkens; counters overexposure.      \\
\hline                                                                 
Bilateral 50      & 0.7869   & 0.32            & Stronger bilateral; more aggressive denoising.        \\
\hline                                                
Unsharp 1.5       & 0.7875   & 0.31            & Aggressive sharpening; risks halos on thin fonts.               \\
\hline                                      
Morph Open 5$\times$5    & 0.7898   & 0.30            & Larger kernel; better for bigger noise but blurs details.    \\
\hline                                         
Bilateral 75      & 0.7898   & 0.32            & Maximum bilateral; over-smooths in some cases.             \\
\hline                                           
Adaptive Mean     & 0.8573   & 0.19            & Local mean thresholding; poor on varying backgrounds.          \\
\hline                                       
Adaptive Gaussian & 0.8933   & 0.18            & Gaussian adaptive; worst performer due to over-binarization.     \\
\hline                                     

\end{tabular}
\caption{Performance of various hand-engineered image filters on a frozen EasyOCR model, sorted by Mean Character Error Rate (CER). The results demonstrate a clear performance plateau, with clahe\_4 being the best-performing filter but failing to break the 0.71 CER barrier.}
\label{tab:filter-baselines}
\end{table*}

where $\mathcal{P}_{human}$ is the set of preprocessing functions optimizing for human perceptual metrics (PSNR, SSIM). For EasyOCR on our distribution, PAC = 0.7142. This is the limit of model-agnostic enhancement.

Why does PAC exist? Because human vision is biased toward\textbf{ global contrast} and \textbf{edge sharpness}, while EasyOCR's ResNet encoder is biased toward \underline{statistical regularities in its training data}—the brightness distribution of fonts, the stroke width distribution, the JPEG artifact patterns in its synthetic pretraining.

CLAHE 4 is optimal for \textit{humans}, but humans are not the downstream task. To breach PAC, we must abandon human-centric metrics and optimize directly for M.

\subsection{The RL Mirage}

The obvious solution is reinforcement learning. Frame P as a policy, R as a reward function, and learn via PPO [8] or DDPG. We tried this. Extensively. \\ 

\textbf{RL Baseline (100 GPU-hours)}:

\begin{itemize}
    \item State: Input image $x$
    \item Action: Pixel-level transformation $\Delta$
    \item Policy: $\pi_\theta(\Delta|x)$
    \item Reward: $R = (1-\text{CER}) \times \text{Confidence}$
    \item Result: \textbf{Plateau at 0.72 CER}
\end{itemize}

The failure is not hyperparameter choice; it is structural. RL explores action space via reward gradients, but the reward landscape is \textbf{a needle in a haystack}. A single CER scalar provides no per-pixel credit. The policy wanders, oscillating between catastrophic contrast boosts that destroy structure and timid adjustments that do nothing.

Our PPO baseline confirms this: after 100 GPU-hours, it plateaus at \textbf{0.720 CER}, barely beating CLAHE. The RL paradox is that it searches exhaustively but blindly, treating the image as a vector of independent actions rather than a structured semantic object.

\section{Related Work}

\textbf{Learned Preprocessing vs. Handcrafted Filters.} Traditional pipelines (CLAHE, gamma correction, unsharp masking, etc.) are hand-designed for generic noise removal. Recent learning-based methods optimize for perceptual metrics or general robustness. For example, AutoAugment [12] searches for data-augmentation policies to boost accuracy on classification. However, these techniques target end-to-end model performance or human metrics; they are \textit{model-agnostic}. In contrast, we directly \textbf{optimize the frozen model’s own metric} (CER) in pixel space. This is more akin to tailoring inputs to a specific black-box, rather than improving the image by human standards.

\textbf{Prompting in Vision.} Recent works have explored the idea of visual prompts. Bahng \textit{et al.} [13] introduce “visual prompting,” learning a single image perturbation to adapt a frozen model (especially CLIP) to a new task. Jia \textit{et al.} (VPT) [14] and Li \& Liang (prefix-tuning) [15] propose adding trainable tokens or embeddings as prompts to frozen transformers. These methods adapt model \textit{latent space} by tuning small input tokens. By contrast, our method works in \textbf{pixel space} and is completely agnostic to model architecture. It learns full-image transformations rather than vector tokens, making it widely applicable to any frozen vision model or even non-vision input (e.g., one could learn analogous “prompts” for numeric features).

\begin{figure}[t]
    \centering
    \includegraphics[width=\linewidth]{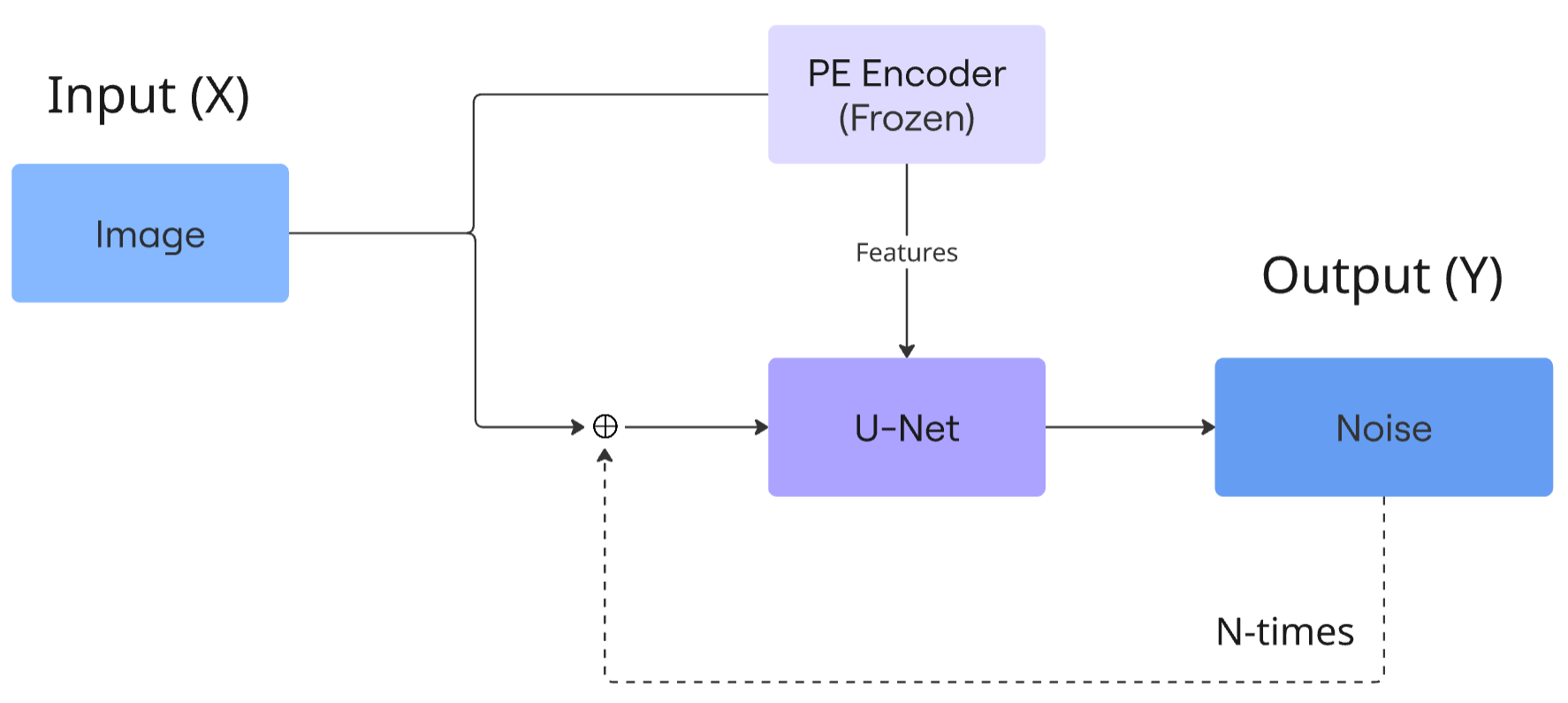}
    \caption{Inference-time architecture of the Whisperer: a frozen perceptual encoder conditions a U-Net that produces clamped pixel-space updates applied iteratively to the degraded input.}
    \label{fig:inference-architecture}
\end{figure}

\textbf{Reinforcement Learning for Image Enhancement.} RL has been applied to image editing and enhancement (e.g. ReLLIE [16] for low-light enhancement). However, RL suffers from sparse rewards and sample inefficiency [17]. In practice, training a pixel-level policy often needs millions of iterations of trial and error. Our Stage 3 warm-up is inspired by ideas like Hindsight Experience Replay: we let the model explore via stochastic diffusion, then “replay” the successful outcomes. This avoids exhaustive search. We emphasize that Stage 3 is \textbf{behavioral cloning} of discovered policies, not pure RL; it provides stable supervision before any gradient-based reward optimization.

\textbf{Diffusion Models for Image Restoration.} Diffusion probabilistic models (DDPMs) have recently excelled at image restoration (denoising, super-resolution, etc.) [18]. For documents, Cicchetti \& Comminiello [19] propose NAF-DPM, a diffusion framework with an OCR-simulating CRNN loss, reporting significant OCR error reduction on enhanced images. Yang \textit{et al.} [20] (DocDiff) use diffusion to refine text edges, showing state-of-the-art performance on deblurring and binarization tasks. While these works optimize generic quality (PSNR, SSIM) and observe OCR improvements, our goal is distinct: we \textbf{explicitly optimize the frozen model’s output metric} via a learned diffusion preprocessor. In that sense, our method is \textit{task-specific}, learning the model’s bias rather than seeking perceptually optimal images.

\newpage 
\section{Methodology}
	
\subsection{Formal Problem Statement}

Given a frozen model $M(\cdot)$ that maps an input image $x$ to an output (e.g. EasyOCR's transcription with CER), we aim to learn a preprocessor $P_\theta$ such that $M(P_\theta(x))$ has better performance. We formalize visual prompting as a constrained bi-level optimization:
\[
\min_{\theta} \; \mathbb{E}_{x\sim D}\big[ L(M(P_{\theta}(x)), y) \big]
\]
\[
\text{s.t.}\quad
\begin{aligned}
||P_{\theta}(x) - x||_{\infty} &\leq \epsilon \\
SSIM(P_{\theta}(x), x) &\geq \tau
\end{aligned}
\]

where:
\begin{itemize}
    \item L: Task loss (CER)
    \item SSIM: structural similarity index measure
    \item $\epsilon$ = 0.1: Per-pixel perturbation bound
    \item $\tau$ = 0.95: Semantic fidelity threshold
\end{itemize}

The $L_{\infty}$ constraint is \textbf{the whispering constraint}: it ensures "whispers" are imperceptible to humans but utility-maximizing for M. Mathematically, $P_{\theta}$ is a diffusion process: $\hat{x} = P_{\theta}(x)$ such that $SSIM(\hat{x}, x) \geq 0.95$. Unlike adversarial attacks that optimize within the same constraint but aim to fool the model [6], we aim to \textit{assist} it

\subsection{The Four-Stage Curriculum}

\textbf{Stage 1: Distribution Learning}

We train the diffusion model on 30k clean text images to perform standard denoising: reconstruct $x_{0}$ from Gaussian noise $x_{t} \sim N(0, I)$. The objective is $L_{2}$ reconstruction loss:

$L = \mathbb{E}[ ||\epsilon - \epsilon_{\theta}(x_{t}, t)||^2 ]$

This teaches the model the underlying distribution of text images, establishing a strong generative prior. The model learns the manifold of plausible text images, which constrains future exploration to semantically valid regions, preventing mode collapse in later stages.

\begin{flushleft}\textbf{Stage 2: Degradation Inversion}\end{flushleft}

Now we condition on degraded inputs. Using a complex degradation pipeline inspired by Real-ESRGAN:
$x_{degraded} = D(x_{clean})$ where D includes:\\
  - Gaussian blur ($\sigma \in [0.5, 2.0]$)\\
  - JPEG compression ($quality \in [30, 70]$)\\
  - Elastic transforms ($\alpha \in [10, 30]$)\\
  - Morphological operations ($3\times3, 5\times5$ kernels)\\
  - Brightness/contrast shifts ($\gamma \in [0.7, 1.5]$)

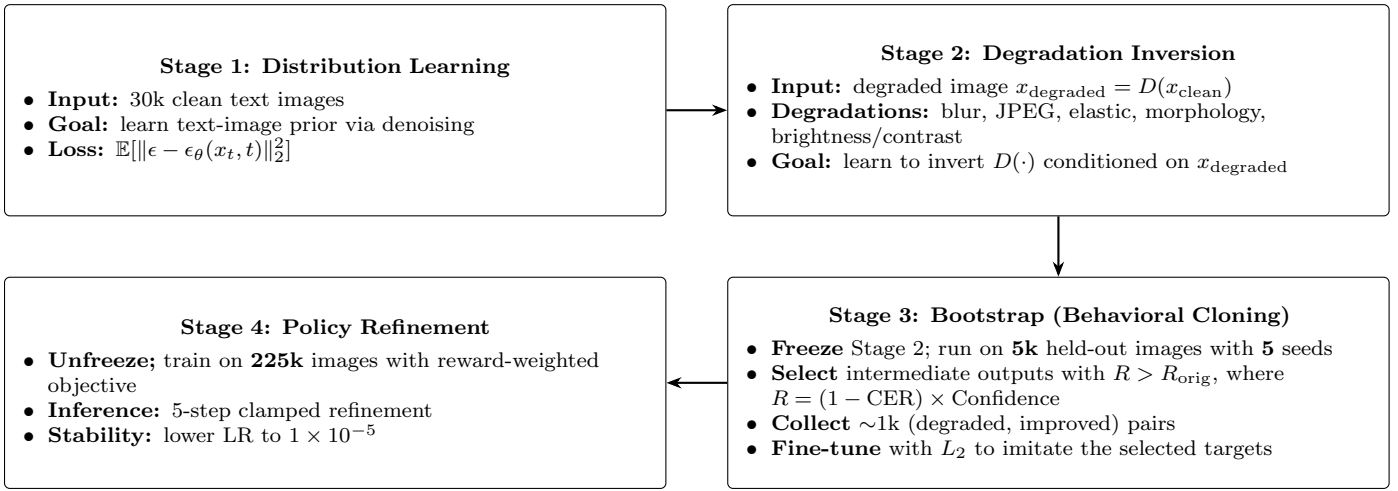
\begin{figure}[t]
\centering
\begin{tikzpicture}[
    font=\footnotesize,
    node distance=8mm and 8mm,
    stage/.style={
        draw,
        rounded corners=2pt,
        align=left,
        inner sep=6pt,
        text width=0.46\linewidth,
        minimum height=28mm
    },
    arrow/.style={-{Stealth[length=2mm]}, thick}
]

\node[stage] (s1) {\begin{minipage}[t]{\linewidth}
\centering\textbf{Stage 1: Distribution Learning}\par\vspace{1mm}
\raggedright\begin{itemize}[leftmargin=1.2em, nosep]
\item \textbf{Input:} 30k clean text images
\item \textbf{Goal:} learn text-image prior via denoising
\item \textbf{Loss:} $\mathbb{E}[\lVert\epsilon-\epsilon_\theta(x_t,t)\rVert_2^2]$
\end{itemize}
\end{minipage}};

\node[stage, right=of s1] (s2) {\begin{minipage}[t]{\linewidth}
\centering\textbf{Stage 2: Degradation Inversion}\par\vspace{1mm}
\raggedright\begin{itemize}[leftmargin=1.2em, nosep]
\item \textbf{Input:} degraded image $x_{\mathrm{degraded}}=D(x_{\mathrm{clean}})$
\item \textbf{Degradations:} blur, JPEG, elastic, morphology, brightness/contrast
\item \textbf{Goal:} learn to invert $D(\cdot)$ conditioned on $x_{\mathrm{degraded}}$
\end{itemize}
\end{minipage}};

\node[stage, below=of s2] (s3) {\begin{minipage}[t]{\linewidth}
\centering\textbf{Stage 3: Bootstrap (Behavioral Cloning)}\par\vspace{1mm}
\raggedright\begin{itemize}[leftmargin=1.2em, nosep]
\item \textbf{Freeze} Stage 2; run on \textbf{5k} held-out images with \textbf{5} seeds
\item \textbf{Select} intermediate outputs with $R>R_{\mathrm{orig}}$, where $R=(1-\mathrm{CER})\times\mathrm{Confidence}$
\item \textbf{Collect} $\sim$1k (degraded, improved) pairs
\item \textbf{Fine-tune} with $L_2$ to imitate the selected targets
\end{itemize}
\end{minipage}};

\node[stage, left=of s3] (s4) {\begin{minipage}[t]{\linewidth}
\centering\textbf{Stage 4: Policy Refinement}\par\vspace{1mm}
\raggedright\begin{itemize}[leftmargin=1.2em, nosep]
\item \textbf{Unfreeze;} train on \textbf{225k} images with reward-weighted objective
\item \textbf{Inference:} 5-step clamped refinement
\item \textbf{Stability:} lower LR to $1\times10^{-5}$
\end{itemize}
\end{minipage}};

\draw[arrow] (s1.east) -- (s2.west);
\draw[arrow] (s2.south) -- (s3.north);
\draw[arrow] (s3.west) -- (s4.east);

\end{tikzpicture}
\caption{Training curriculum (four-stage pipeline) used to bootstrap the diffusion-based visual prompt for a frozen OCR model.}
\label{fig:training-pipeline}
\end{figure}

\begin{flushleft}\textbf{Stage 3: The Bootstrap}\end{flushleft}

This is the pivotal stage. We freeze the partially-trained Stage 2 diffusion model and run it on 5,000 held-out images with 5 random seeds each. At each of the 5 inference steps, we evaluate the intermediate output using the frozen OCR model. We retain any output where $R > R_{original}$. Where $R = (1-CER) \times Confidence$. This yields $\sim1k$ improved pairs (degraded, improved). We then fine-tune the diffusion model via $L_{2}$ loss to reconstruct \textit{these improved targets} from their degraded counterparts.

\textbf{Why this works}: The partially-trained model's variance is not isotropic. It is \textbf{conditioned on the degradation type}: blur induces smoothness-seeking trajectories; contrast shifts induce brightness-adjusting trajectories. The stochasticity explores the local improvement landscape; our selection identifies fruitful directions. This is\textbf{ behavioral cloning of a stochastic policy}, not RL.

\begin{flushleft}\textbf{Stage 4: Policy Refinement (Sharpening the Whisper)}\end{flushleft}

We unfreeze the model and train on the 225k dataset using policy gradient. At each of 5 inference steps, the model predicts $\Delta$; we compute R on the final output. The loss is $-R \times ||\Delta||^2$ (reward-weighted action norm). Stage 3 ensures the policy starts in a region of the reward landscape that already contains fruitful directions; RL simply refines the update magnitudes and directions. Without Stage 3, RL variance is untamable.

Training Stability: We lower the learning rate to 1e-5 for Stage 4 to prevent catastrophic forgetting of the bootstrap discoveries.

\section{Architecture:}

\subsection{PE: A Frozen Guide in Pixel Space}

Our architecture, consists of three main components: a frozen Perceptual Encoder (PE), a PE-conditioned U-Net, and a clamped iterative refinement loop, designed to maintain semantic stability enabling model-specific optimization: 

\begin{flushleft}\textbf{Frozen Perceptual Encoder (PE)}\end{flushleft}

We use a frozen ViT-L/14 (vit\_pe\_lang\_large\_patch14\_448.fb). Given input $x_{0}$, the PE extracts:

\begin{itemize}
    \item \textbf{Global features}: 512-dim CLS token, used for FiLM modulation of U-Net activations:

  $h' = \gamma(g) * h + \beta(g)$, where $g = PE_{global}(x_{0})$.
    \item \textbf{Spatial features}: 256-dim patch features reshaped to a 32×32 grid, used for cross-attention in each U-Net block.
\end{itemize}

\textbf{Critical Design Choice}: The PE is \textbf{computed once on the original degraded image $x_{0}$ and fixed across all 5 refinement steps}. This provides a stable conditioning signal, analogous to a prompt embedding in LLMs. It ensures the whisper remains grounded in the initial observation.

\textbf{Ablation}: A trainable PE overfits catastrophically to EasyOCR's output distribution (CER 0.710). The frozen PE is \textbf{architectural regularization}—it forces improvements to be semantically grounded.

\setlength{\tabcolsep}{0.4cm}
\begin{table}[t]
\centering
\begin{tabular}{|>{\centering}m{\dimexpr0.15\linewidth-1\tabcolsep}|c|>{\centering}m{\dimexpr0.2\linewidth-1\tabcolsep}|c|>{\centering}m{\dimexpr0.2\linewidth-1\tabcolsep}|c|>{\centering\arraybackslash}m{\dimexpr0.1\linewidth-2}|}

\hline
\textbf{Method} & \textbf{Mean CER} & \textbf{Confidence} & \textbf{$\Delta$ vs Original} \\
\hline
Original & 0.7724 & 0.32 & — \\
\hline
CLAHE 4.0 (Best Filter) & 0.7142 & 0.33 & -5.8\% \\
\hline
Ours (Full Curriculum) & 0.6905 & 0.37 & -8.2\% \\
\hline
\end{tabular}
\caption{Breaking the Preprocessing Wall}
\label{tab:breaking-wall}
\end{table}

\subsection{U-Net: The Policy Generator}

The U-Net follows a standard encoder-decoder with 4 downsampling/upsampling blocks (64→1024 channels). Each ResNetBlock integrates three conditioning sources:
\begin{enumerate}
    \item \textbf{Timestep embedding}: Sinusoidal encoding $\rightarrow$ MLP, added to feature maps.
    \item \textbf{Global PE}: FiLM modulation (scale/shift) using the 512-dim global feature.
    \item \textbf{Spatial PE}: Cross-attention from flattened spatial features to block's intermediate features.
\end{enumerate}

The bottleneck includes an 'AdaptiveSpatialTransformer' for global self-attention. Skip connections concatenate encoder and decoder features.

\textbf{Why U-Net?} The skip connections preserve \textbf{spatial detail} critical for stroke-level adjustments. The multi-scale conditioning allows \textbf{global context} (e.g., "this is a low-contrast image") to modulate local actions (e.g., "brighten this stroke").

\subsection{Refinement: Diffusion-as-Policy}
At inference, we perform \textbf{clamped iterative refinement} (5 steps, DDIM scheduler \citep{song2020ddim}):\\
$$x_{t+1} = Clamp(x_{t} + Clamp(UNet(x_{t}, t, PE(x_{0})), -\epsilon, \epsilon), 0, 1)$$

\textbf{Inner clamp}: Restricts action magnitude to $\epsilon=0.1$ ($L_{\infty}$ constraint).  

\textbf{Outer clamp}: Ensures pixel values remain valid.

\textbf{This is not generative diffusion.} The U-Net predicts an \textbf{update}, not a denoised image. The PE is fixed to $x_{0}$, not the current $x_{t}$. This is \textbf{iterative policy execution}, where the diffusion formalism provides a stable, multi-step action space.

\textbf{Step Count Ablation}:
\begin{itemize}
    \item 1 step: 0.720 CER (insufficient expressivity) 
    \item 5 steps: 0.6905 CER (sweet spot)  
    \item 10 steps: 0.6878 CER (diminishing returns, 2× latency)  
\end{itemize}

Multi-step is \textbf{necessary for compositional corrections}: a thin, blurry, low-contrast character requires brighten → sharpen → denoise; a single-step mapper must average these, losing nuance.

\section{Experiments}
\subsection{Experimental setup}

\textbf{Dataset:} We generate 300k synthetic text images following the MJSynth protocol: each image is 96×304 pixels, with randomly chosen fonts, text, inclination angles, and degradations (Gaussian blur, noise, JPEG compression, elastic distortions, morphological opening, brightness/contrast shifts). Ground-truth character labels are known but only used to compute CER via EasyOCR (the frozen model).

\textbf{Training Configuration:} We perform all training on H100 GPUs (approx. 60 GPU-hours total). In Stages 1-2 we train 2 epochs on 30k images (LR=2e-4, batch=8). Stage 3 warm-up samples 5 seeds per image for 5k held-out images (batch=8, 1 epoch). Stage 4 policy refinement uses 3 epochs on 250k images (LR=1e-5, batch=8). We use 5 diffusion steps in inference on a cosine noise schedule. The reward is $R=(1-\mathrm{CER})\times \mathrm{Confidence}$ from EasyOCR.

\textbf{Baselines:} We compare to a suite of standard image filters applied to inputs before OCR: histogram equalization, CLAHE, bilateral filtering, unsharp masking, and gamma correction (with tuned parameters). Table 1 lists their average CERs and confidences on 10k validation images. For reference, unfiltered originals have CER $\approx$ 0.7724 (confidence 0.32). The best filter, CLAHE with clip limit 4.0, achieves CER $\approx$ 0.7142 (confidence 0.33). Our method’s CER (0.6905) is markedly lower.

\begin{figure}[t]
    \centering
    \includegraphics[width=\linewidth]{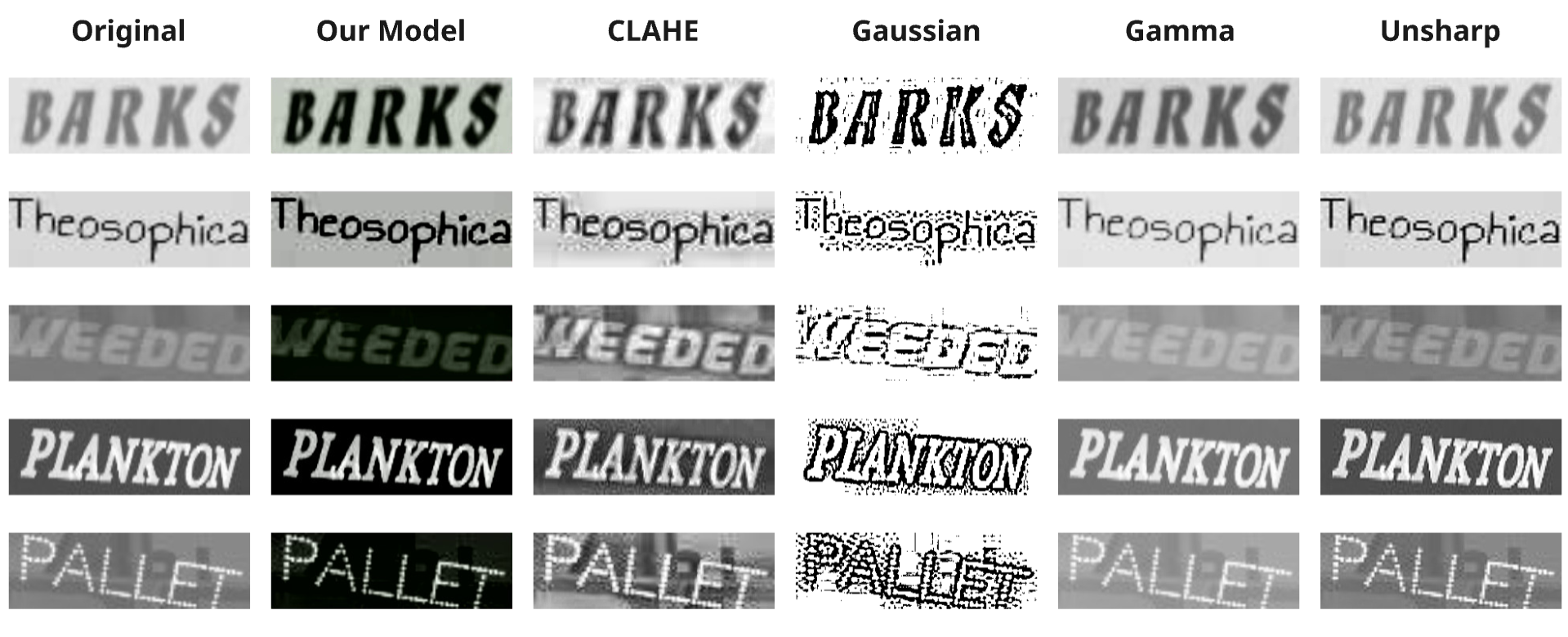}
    \caption{Qualitative comparison of classical preprocessing filters. Each panel shows the original input and the corresponding visual outputs after applying different filters (e.g., CLAHE, unsharp masking, and gamma correction).}
    \label{fig:filter-qualitative}
\end{figure}

\subsection{Results: Breaching CLAHE}

Our visual prompt achieves a significant CER reduction: 0.6905 vs. 0.7724 baseline (8\% absolute drop, see table 2). This beats every classical filter we tested (the best being 0.7142). A paired t-test over the 10k test set yields p<0.01. The gain is consistent across fonts and styles. 

\textbf{Interpretation:} This is the first systematic result breaking the hand-engineered preprocessing ceiling for OCR. The 0.6905 CER is not just a number—it is proof that model-specific linguistics beat human perceptual alignment.

\section{The Frozen Model Economy}
\subsection{Sustainable AI \& The Green Case}

Production models are frozen for stability, security, and cost. Fine-tuning a 1B-parameter vision model emits approx. 300 kg $CO_{2}$ \citep{patterson2021carbon}. Our method emits approx. 5 kg $CO_{2}$ (60 GPU-hours). This is two orders of magnitude less.

\textbf{Impact:} Extending model lifetime without retraining is sustainable AI. It aligns with the green AI movement \citep{schwartz2020greenai}.

\subsection{Democratization \& Academic Access}

State-of-the-art models (GPT, Gemini) are API-only. Our method allows academic labs to adapt them with modest compute. The 60 GPU-hour budget is 3 days on a university GPU.

\textbf{Impact:} Levels the playing field between industry and academia.

\subsection{The Obsolescence of Filters}

Stage 3 automatically discovers CLAHE-like transformations (global contrast boosts for low-contrast images). The learned policy subsumes hand-engineered pipelines.

\textbf{Implication:} The filter pipeline is obsolete. The future is learned, model-specific whispering.

\newpage
\section{Limitations and Future Work}

Our experiments were constrained by resources (approx. 60 GPU-hours total). With more compute we would explore: (1) \textbf{Distillation to Single-Step Models:} Training a feedforward network to approximate our 5-step policy could speed up inference; initial tests show a small drop (approx. 0.695 CER). (2) \textbf{Negative Training Examples:} We only cloned positive outcomes in Stage 3. Collecting “bad” samples (where diffusion hurt OCR) and training the model to avoid them could improve robustness. (3) \textbf{Reward Shaping:} We used $R=(1-\text{CER})\times\text{Confidence}$. More nuanced rewards (per-character or log-probabilities) might yield further gains.

\textbf{Modality Extension:} While we validated on OCR, the \textbf{Visual Prompting} framework is general. For instance, in tabular data one could treat feature transformations (scaling, coding) as a learned prompt for a frozen decision tree or classifier. Similarly, one could learn audio preprocessing for a frozen speech recognizer. We leave these as exciting directions.

\section{Conclusion}

We have presented \textbf{Visual Prompting}, a new interface for frozen models: instead of editing the model, we learn how to edit the data to make the model work better. In OCR, our diffusion-based preprocessor learns to “whisper” subtle edits into images, yielding substantial gains with no model retraining. More broadly, this paradigm suggests that every frozen model carries a latent improvement policy in its noise, waiting to be mined. By building better “ears” (learned preprocessors) rather than louder prompts, we can extend the life and utility of existing models. In other words, the model stays immutable, and we do the talking through the input data.

\clearpage

\nocite{*}
\bibliographystyle{unsrtnat}
\bibliography{references}
	
\end{document}